\documentclass[screen,authorversion]{acmart}     

\usepackage[utf8]{inputenc}        
\usepackage[nameinlink]{cleveref}  
\usepackage{colortbl}              
\usepackage{soul}                  
\usepackage{fontawesome5}          

\newcommand{\enquote}[1]{``#1''}  

\newcommand{\laptop}{\scriptsize\faLaptop}
\newcommand{\workstation}{\scriptsize\faDesktop}
\newcommand{\cluster}{\scriptsize\faServer}

\copyrightyear{2025}
\acmYear{2025}
\setcopyright{rightsretained}
\acmConference[LAK 2025]{LAK25: The 15th International Learning Analytics and Knowledge Conference}{March 03--07, 2025}{Dublin, Ireland}
\acmBooktitle{LAK25: The 15th International Learning Analytics and Knowledge Conference (LAK 2025), March 03--07, 2025, Dublin, Ireland}
\acmDOI{10.1145/3706468.3706485}
\acmISBN{979-8-4007-0701-8/25/03}

\begin{document}

\title{Evaluating the Impact of Data Augmentation on Predictive~Model~Performance}

\author{Valdemar Švábenský}
\orcid{0000-0001-8546-280X}
\affiliation{
    \institution{Kyushu University}
    \city{Fukuoka}
    \country{Japan}
}
\email{valdemar.research@gmail.com}

\author{Conrad Borchers}
\orcid{0000-0003-3437-8979}
\affiliation{
    \institution{Carnegie Mellon University}
    \city{Pittsburgh, PA}
    \country{USA}
}
\email{cborcher@cs.cmu.edu}

\author{Elizabeth B. Cloude}
\orcid{0000-0002-7599-6768}
\affiliation{
    \institution{Michigan State University}
    \city{Lansing, MI}
    \country{USA}
}
\email{ecloude94@gmail.com}

\author{Atsushi Shimada}
\orcid{0000-0002-3635-9336}
\affiliation{
    \institution{Kyushu University}
    \city{Fukuoka}
    \country{Japan}
}
\email{atsushi@ait.kyushu-u.ac.jp}


\begin{abstract}
In supervised machine learning (SML) research, large training datasets are essential for valid results. However, obtaining primary data in learning analytics (LA) is challenging. Data augmentation can address this by expanding and diversifying data, though its use in LA remains underexplored. This paper systematically compares data augmentation techniques and their impact on prediction performance in a typical LA task: prediction of academic outcomes. Augmentation is demonstrated on four SML models, which we successfully replicated from a previous LAK study based on AUC values. Among 21 augmentation techniques, SMOTE-ENN sampling performed the best, improving the average AUC by 0.01 and approximately halving the training time compared to the baseline models. In addition, we compared 99 combinations of chaining 21 techniques, and found minor, although statistically significant, improvements across models when adding noise to SMOTE-ENN (+0.014). Notably, some augmentation techniques significantly lowered predictive performance or increased performance fluctuation related to random chance. This paper's contribution is twofold. Primarily, our empirical findings show that sampling techniques provide the most statistically reliable performance improvements for LA applications of SML, and are computationally more efficient than deep generation methods with complex hyperparameter settings. Second, the LA community may benefit from validating a recent study through independent replication.
\end{abstract}



\maketitle

\section{Introduction}

To improve teaching and learning, primary student data (such as data on students' learning progression, behavior, and outcomes) are essential for studying scientific questions in learning analytics (LA) research. High quality and adequate volume of research data are especially important when the research methods involve machine learning (ML), such as predictive modeling, which is extensively used in LA~\cite{sghir2023recent, gavsevic2017piecing}. However, collecting such data is challenging due to several reasons. First, it is time-intensive and costly to capture data from a sufficiently large sample of participants. In addition, the data samples usually represent homogeneous learner populations, meaning that they are bound to a specific context, time, and learning system~\cite{Baker2019}. As a result, datasets can often be small and have low diversity, limiting predictive model performance and generalizability. Second, ensuring data privacy requires thorough anonymization, and while tools are being developed to help automate this task~\cite{Singhal2024deidentifying}, it is still time-consuming and error-prone. Third, additional ethical and legal issues are associated with collecting, storing, and processing student data~\cite{hutt2023right, klose2020edm}. This is especially true for sensitive data collected in schools or other research settings that can personally identify a student. Prominent examples of such data include students' writing style or speech, for example, in discussion forum posts~\cite{Henricks2024}, student discourse~\cite{Dang2024}, or written peer feedback on problem-solving~\cite{Hutt2024}.

\subsection{Data Augmentation}

There is opportunity to address the aforementioned challenges of many datasets' limitations in terms of size and diversity by utilizing \textit{data augmentation}: a set of techniques~\cite{Maharana2022augmentation} to \enquote{increase the volume, quality and diversity of training data}~\cite{Mumuni2022augmentation}, while simultaneously maintaining students' privacy and adhering to ethical standards of research. Data augmentation involves computational methods for sampling new data, transforming existing data, and generating synthetic data~\cite{yue2018synthetic} (see details in \Cref{subsec:related-work-rq2}). At their core, these methods enhance the predictive performance of ML models by leveraging underutilized signals in imbalanced data or rare data patterns. At the same time, they regularize the model through noise addition, akin to other regularization techniques like dropout or weight penalties. 

These techniques have proven effective for improving ML model performance and generalizability in data science applications across various domains~\cite{Chui2024-augmentation-survey}, such as image processing~\cite{Shorten2019augmentation}, computer vision~\cite{Jaipuria2020deflating}, and healthcare~\cite{Pezoulas2024augmentation}. However, the utility of data augmentation depends on dataset characteristics (e.g., size, structure, sparsity)~\cite{Bayer2022,Shorten2019augmentation}. In non-LA domains~\cite{Bayer2022,Shorten2019augmentation,Jaipuria2020deflating,Pezoulas2024augmentation}, augmented datasets often have properties different from typical LA datasets. Since augmentation is underexplored within LA-relevant datasets, much is still unknown about its effect and influence on ML model performance in LA research. Therefore, its viability must be rigorously evaluated to improve critical LA tasks. 

Data augmentation, especially synthetic data generation, is closely related to simulated data. Recently, \citet{Kaser2024simulated} published \enquote{the first systematic literature review on simulated learners}, which uncovered challenges and limitations for current simulated LA datasets. First, they argued that \enquote{simulated learner models tend to represent only narrow aspects of student learning}, as most prior work focused on simulating a small set of cognitive skills~\cite{Kaser2024simulated}. Yet, non-cognitive factors are also essential to learning and influence student outcomes~\cite{almeda2020predicting}. Thus, learner models should represent multiple facets of learning guided by theoretical perspectives on human knowledge and skill acquisition. Second, several studies do not provide evidence on the validity of simulated learner models and downstream findings generated from them. More research is needed to address these limitations. This is one of the aims of the current study, which uses data augmentation to improve model predictive validity in a common LA classification task.

\subsection{Research Reproducibility and Replicability}
\label{subsec:intro-repr}

An additional aspect that improves the validity of learner models and the ability to draw generalizable conclusions is the successful reproduction and replication of research findings. According to the Association for Computing Machinery (ACM), \textit{reproducibility} of research means that a result from a computational experiment can be independently obtained by a different team (other than the original authors) using data and artifacts from the original author team~\cite{acm-repr}. This is a key property of good science, as it ensures credible results and enables their broader applicability. A term related to reproducibility is \textit{replicability}, which means that a result can be obtained by a different team using different artifacts~\cite{acm-repr}. Currently, very few published LA studies have been replicated, which we describe more thoroughly in \Cref{subsec:related-work-rq1}. Moreover, AI research as a whole (which includes machine learning in LA), also faces a reproducibility crisis~\cite{Hutson2018crisis}.

\subsection{Goals and Scope of This Paper}
\label{subsec:intro-goals}

Our paper uniquely addresses both replication and data augmentation in the LA context. Specifically, we augment existing student data selected from studies recently published at the LAK conference. Since these studies have already passed the peer review process, the existing models should have higher validity than if we developed new models from scratch. To narrow our focus, we choose the context of predictive ML modeling, which is central to LA research~\cite{sghir2023recent, gavsevic2017piecing}. ML is a leading topic especially at conferences like LAK due to its implications for building personalized and adaptive systems that guide just-in-time interventions, such as dropout prediction~\cite{sghir2023recent}, knowledge tracing~\cite{am2021literature}, affect detection~\cite{kai2015comparison}, and the evaluation of predictive analytics interventions in the classroom~\cite{handbook-la2017}. However, valid ML research requires robust models and generalizable findings~\cite{Baker2019}, which can be supported by replication and data augmentation. 

Focusing on research that reported predictive ML models, we first attempt to \textit{replicate} these models and the corresponding findings. Second, to evaluate the utility of data augmentation, we \textit{re-train} these models with a mixed dataset consisting of original and augmented data. 
We hypothesize that the re-trained ML models will demonstrate better performance than the original ML models due to better quality and more diverse data for training.

Within this scope, our paper investigates the following research questions (RQs):
\begin{enumerate}
    \item[RQ1] \textit{To what extent can we replicate the analyses and results from a selected previous learning analytics study?}
    \item[RQ2] \textit{When data augmentation techniques are applied to the original training data, which of these techniques, and to what extent, improve model performance?}
\end{enumerate}

\subsection{Contributions}

Our paper is original in tackling both the issues of replication and data augmentation in parallel. Combining replication and augmentation is valuable for two reasons. First, modeling methods were validated through prior peer-review, ensuring they meet LAK's methodological community standards (e.g., cross-validation). Second, replicating published models establishes a competitive baseline for demonstrating reliable performance improvements through augmentation. 

Our paper's first contribution is replicating an LA study that predicts long-term academic outcomes using ML models with features grounded in a learning theory.
Second, we use the replicated models as a case study to demonstrate and evaluate whether augmentation techniques improve prediction, and in which cases. We measure changes in model performance to systematically compare whether different augmentation techniques yield statistically significant improvements.
Third, we offer a methodological contribution in showing how to evaluate augmentation techniques and offer practical recommendations and lessons learned. All code is available for other researchers to adopt or adapt (see \Cref{sec:conclusion}). This enables researchers to augment their own datasets and study the significance of prediction performance improvements in relationship to a suite of augmentation techniques for their own educational models.

From a high-level perspective of the four phases of the LA cycle~\cite{Clow2012cycle}, our research focuses on improving \textit{data} to build more robust, valid, and generalizable models during the \textit{metrics} phase. As a result, improving ML models holds implications for yielding more accurate educational \textit{interventions} to better support \textit{learners}. Lastly, it contributes to creating more effective personalized systems that can better support teaching and learning practices.

\section{Related Work and Novelty of This Paper}
\label{sec:related-work}

We review prior work for each RQ: \Cref{subsec:related-work-rq1} focuses on replication, and \Cref{subsec:related-work-rq2} on data augmentation. In both sections, the last paragraph highlights the novelty of our paper. In addition, \Cref{subsec:related-work-theory} introduces theory that grounded our approach -- another aspect that has been insufficiently covered in literature~\cite{Kaser2024simulated} and is addressed by our paper.

\subsection{RQ1: Research Reproducibility and Replicability in LA}
\label{subsec:related-work-rq1}

In the ACM Digital Library, which indexes all past 14 years of LAK conference publications, we searched for terms related to reproducibility and replicability (search query: \texttt{reproduc* OR replica*}) in the titles, abstracts, or keywords of the published articles. Constraining the search to the most recent five years of LAK (2020--2024) yielded nine papers, out of which seven performed reproduction or replication as defined in \Cref{subsec:intro-repr}. Within this five-year search period, a total of 379 full and short papers were published at LAK, which means that only 1.8\% included an element of reproducing/replicating previous research. Moreover, out of these 1.8\% that replicated past research results, 0\% performed data augmentation with the aim of improving predictive performance, as explored in our present study.

The finding that only seven of the 379 (1.8\%) LAK papers in the past five years reproduced/replicated a past study can partially be attributed to the lack of reproducibility described by \citet{Haim2023lak}. They estimated the reproducibility of all LAK 2021 and 2022 papers, investigating whether the papers' data and supplementary materials (e.g., code) were documented well, such that another researcher could (with reasonable effort) reproduce the findings. Only 5\% of papers made their raw dataset available, and in another 2\%, data could be requested from the authors, effectively making at least 93\% of LAK papers \textit{not reproducible}. In the related field of educational data mining (EDM), the ratios of data sharing practices were slightly higher, but still low: 15\% available and 5\% on request~\cite{Haim2023edm}, making 80\% of papers unreproducible. 

Ultimately, \citet{Haim2023lak} marked all LAK 2021--2022 papers as \textit{not reproducible} within the time slot they allocated for each paper, but \enquote{estimated that the 2\% of papers that contain both the raw dataset and source [code] were likely to be reproducible}. This is a surprisingly low number, given that datasets and computational methods are central to the progress of LA research. Despite efforts that strongly recommend and promote the sharing of data and code among researchers~\cite{Kathawalla2021easing}, it is still not a prevalent practice. Since reproducing research is so rare, this creates a gap in the LA literature, which may lead to questioning the validity and generalizability of some LA results. Our paper contributes towards addressing this gap by validating a recently published study through independent replication.

\subsection{RQ2: Data Augmentation and Its Use in LA}
\label{subsec:related-work-rq2}

In line with our goals (see \Cref{subsec:intro-goals}), we surveyed data augmentation approaches used for improving predictive performance of ML models. We distinguish between \textit{sampling}, \textit{perturbation}, and \textit{generation}. Within these categories, the individual techniques are often \textit{symbolic} (using statistical models) or \textit{rule-based} (using programs or templates)~\cite{Maharana2022augmentation}.

\textit{Sampling methods} create additional data points by resampling from an existing dataset. Common approaches in classification contexts include oversampling, where minority class instances are duplicated, and undersampling, where majority class instances are reduced. A prominent example is the Synthetic Minority Over-sampling Technique (SMOTE), which creates synthetic samples for the minority class by interpolating between existing observations \cite{chawla2002smote}. Adaptive Synthetic Sampling (ADASYN) improves oversampling by creating more samples in regions with sparser data distribution, thereby adapting to the difficulty of learning in different regions of the feature space \cite{he2008adasyn}. Both SMOTE and ADASYN slightly improved the accuracy of student performance prediction in a prior LA study~\cite{ashfaq2020managing}.

\textit{Perturbation methods} (also called \textit{data transformation}~\cite{Mumuni2022augmentation}) involve making small, controlled changes to the original data. They modify the existing training samples or the derived features to increase variance in the feature space during model training. Representative techniques such as noise injection add random values to existing samples, enhancing the models' generalization by preventing overfitting to the original data \cite{shen2022data}. \citet{wei2019eda} proposed four augmentation techniques using random insertion, swapping, and deletion of words in text classification, boosting test set performance on small datasets. To the best of our knowledge, no past work has systematically investigated perturbation in LA.

\textit{Generation methods} (also called \textit{data synthesis}~\cite{Mumuni2022augmentation}) for creating new data have diverse LA applications: from data anonymization, through benchmarking~\cite{berg2016role}, to improving ML models~\cite{zhang2024_3dg}. Various learning processes and outcomes have been studied using synthetic data~\cite{berg2016role, Zhan2023} such as tabular data~\cite{Liu2024}, interaction data~\cite{Flanagan2022}, simulated learner models~\cite{Kaser2024simulated}, or biased data~\cite{Jiang2024}. GPT-4 produced a synthetic dataset of a student-tutor conversation~\cite{Sonkar2024}. Next, generative adversarial networks (GANs)
are especially useful for imbalanced data as they can oversample the minority class \cite{biswas2023generative}. \citet{zhang2024_3dg} used GANs to augment log data from tutoring systems, improving the reliability of assessing student knowledge estimates using the augmented samples. This reliability was not improved when GPT-4 was used.

In past research, the validity of synthetic data has often not been systematically studied~\cite{Kaser2024simulated}, raising the question of whether synthetic data accurately represent real learner data. This is crucial because valid measures for assessment and outcome prediction are central LA goals \cite{divjak2023assessment}. A common approach for determining validity is \textit{predictive validity}: a measure is valid if it accurately predicts outcomes of interest~\cite[Chapter 3]{handbook-la2017}. For example, standardized test scores are a valid representation of academic aptitude if they can predict students' future academic success. In line with this definition, our study addresses the lack of prior validation of synthetic data and augmentation methods. We compare the performance of models predicting long-term academic outcomes, with and without augmentation methods applied.

\subsection{Learning Theory and Theoretical Grounding for RQ1 and RQ2}
\label{subsec:related-work-theory}

Only 3\% of simulated learner models in LA and related fields were anchored in theories of human learning~\cite{Kaser2024simulated}. Moreover, simulated learning often overemphasizes cognitive aspects~\cite{maclellan2016apprentice}, omitting non-cognitive facets of learning such as affective states (e.g., confusion, boredom, engaged concentration). However, to simulate realistic and holistic learner models, it is essential to consider these non-cognitive facets as well, since they influence outcomes and achievement~\cite{d2012dynamics}. 

We ground our RQs in the \textit{Model of Affective Dynamics}~\cite{d2012dynamics} (MAD), a prominent contemporary theory that describes human learning. It explains how cognition, knowledge, and affect interact as learners engage with digital learning systems. As learners assimilate new information into their existing knowledge, they often encounter impasses -- discrepancies between new information and prior understanding -- leading to confusion. Whether this confusion is resolved determines the transition into different affective states, some of which hinder learning. If learners cannot overcome the impasse, confusion may escalate into frustration, and if it persists, further into boredom and disengagement. Conversely, if the impasse is resolved, learners return to a state of engaged concentration, which is conducive to learning.

Empirical evidence shows that learners' affective states significantly influence their long-term academic outcomes, such as college enrollment~\cite{pedro2013predicting}. Outcomes are positively correlated with engaged concentration~\cite{almeda2020predicting} and negatively with boredom and disengagement (e.g., gaming the system)~\cite{mogessie2020confrustion}.
However, \citet{karumbaiah2021re}'s findings contradict MAD, implying that individual factors (such as the learner's knowledge) influence how affective states interact with cognition~\cite{cloude2022affective}, impacting outcomes differently. Specifically, the impact of boredom, confusion, and frustration remains inconclusive: some studies suggest these states are beneficial, while others find them harmful~\cite{d2014confusion,richey2019more,baker2010better,pardos2014affective}. These inconsistencies could be explained by insufficient or low-quality data, which our study attempts to address.

\section{Research Methods}
\label{sec:methods}

\begin{figure*}[!ht]
    \centering
    \includegraphics[width=0.55\textwidth]{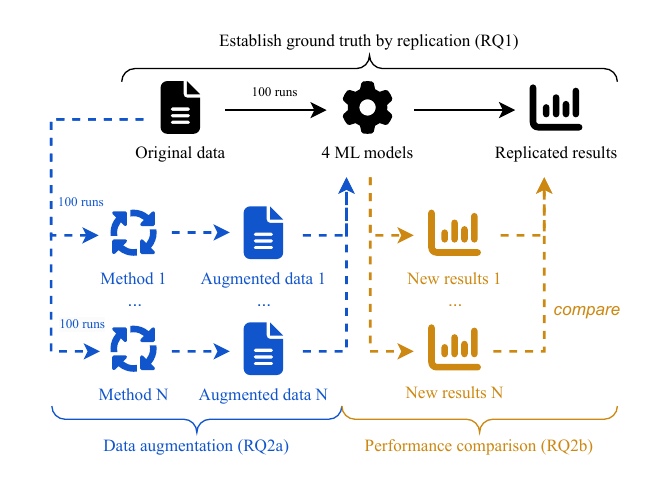} 
    \caption{Conceptual illustration of the research methods and steps, along with the relationship between the two RQs.}
    \Description{We first establish ground truth by replicating the original models. Then, we systematically evaluate several data augmentation methods in a uniform way, and compare the performance of the models trained on the augmented data with the baseline models.}
    \label{fig:method}
\end{figure*}

The goal of the present research is to replicate an LA study that involves predicting an academic outcome, and subsequently investigate whether data augmentation boosts model performance on the same task. Initially, we defined criteria for selecting a study for replication (see \Cref{subsec:methods-criteria}). Next, \Cref{subsec:methods-selected-paper} describes the criteria application and relevant aspects of the selected LA study. \Cref{subsec:methods-rq1} details our methods for replicating the study (RQ1). Finally, \Cref{subsec:methods-rq2} proposes and justifies the methods for data augmentation (RQ2). \Cref{fig:method} illustrates our overall approach.

\subsection{Defining and Justifying the Criteria for Selecting a Paper to Replicate}
\label{subsec:methods-criteria}

To choose a study for replication (RQ1), the research team discussed criteria that a paper should satisfy to attempt replication. When it comes to research data and methods documentation, we drew inspiration from \citet{Haim2023lak}, who systematically studied replication in LA. After several iterations and refinement, we defined that the study must:
\begin{enumerate}
    \item Be a full or short paper published at the \textit{LAK conference within the past five years} (i.e., from 2020 to 2024).
    \item Have \textit{no overlap of co-authors} with the team of our paper, to perform true replication by a different team.
    \item Have research data that are:
    \begin{enumerate}
        \item \textit{Open}, which we define -- in accordance with \citet{Haim2023lak} -- as available either through a free online download, or on request with a latency of less than one month for anyone in the broader research community.
        \item \textit{In English language} for all of the associated results, as there are some LAK 2020--2024 papers with data in Portuguese or Chinese, enabling the data reuse and interpretation only to researchers fluent in these languages.
        \item \textit{Not simulated/synthetic}, since one of the goals of our paper is to determine improvements of augmenting real-world learner data with synthetic or otherwise modified data (RQ2).
    \end{enumerate}
    \item Have research methods that are \textit{documented}, which we define -- in line with \citet{Haim2023lak} -- as allowing us to reproduce or replicate the original study's results (e.g., key outcome measures) within reasonable time.
    \item Have an analytical research objective that involves \textit{prediction by using supervised machine learning} (SML), a central application area of LA \cite{gavsevic2017piecing}. This is because our RQ2 aims to evaluate augmentation techniques, which are typically used for improving the prediction of outcomes of interest through augmented predictors \cite{ashfaq2020managing,shen2022data,wei2019eda,biswas2023generative}.
    \begin{enumerate}
        \item The relationship between the outcome variable and the predictors must be grounded in a learning theory, involving learning process features and academic outcomes derived from educational data.
        \item The SML models must use a test set for evaluation. This allows for reliably estimating predictive performance improvements achieved by data augmentation applied to the training data and tested on the separate test set.
        \item The modeling methods must follow educational data science research standards~\cite{handbook-la2017, handbook-edm2010}, including cross-validation and a comparison of several prediction schemes with different architectural complexity. This ensures any improvements to model performance can be attributed to augmentation, not lack of methodological rigor.
    \end{enumerate}
\end{enumerate}

\subsection{Selected Paper for Replication}
\label{subsec:methods-selected-paper}

At the beginning, 379 candidate papers satisfied criterion (1), as also shown in \Cref{subsec:related-work-rq1}. Applying criterion (2) left us with 371 papers.
Out of these, 36 had open research data based on criterion (3a), 
which is 9.7\%. This ratio of data sharing is similar to, though slightly higher than, the 7\% reported by~\citet{Haim2023lak}. Applying the rest of criterion (3), we were left with 31 papers.
Out of these, 18 papers partially satisfied (4), although some rather minimally, and two papers satisfied (5). For one of the two papers, upon more thorough inspection we realized that we needed additional data from the authors to replicate the findings, but they did not reply to our email and thus the paper was discarded.

We ultimately selected the study by~\citet{zambrano2024long}, which investigated whether students' topic-level knowledge in mathematics along with learning behaviors could predict two outcomes: \textit{college enrollment} and \textit{STEM career choice}. This study was grounded in a theoretical framework of human learning~\cite{d2012dynamics} and discussed the empirical evidence that justified the selection of features used in the ML models.
The study topic is directly related to the key goals of LA, which center around generating insights to support teaching, learning, and educational management~\cite{handbook-la2017}.

\subsubsection{Dataset}
The study used a public dataset~\cite{Patikorn2020} from ASSISTments: a digital learning platform designed to improve mathematical skills~\cite{Heffernan2014}. There were 1,709 students from four middle schools in the USA, who practiced tasks grouped into 12 topic areas~\cite[Table 1]{zambrano2024long}, such as algebra, geometry, and functions.
This dataset tracked students longitudinally: starting from their interactions with ASSISTments and behaviors during learning activities, to their college enrollment and their first job (STEM vs. non-STEM). We obtained free access to the dataset after filling in the request form~\cite{Patikorn2020}. 

\subsubsection{Target Variable}
We decided to focus only on the \textit{enrollment} outcome and not \textit{STEM career}. This is because for predicting \textit{enrollment}, the paper used the full dataset of 1,709 students, but for \textit{STEM career}, only about a third of the dataset (591 students) included a coded outcome. This small sample size would make it difficult to establish significantly different ML model performance results between baseline and augmented models in our RQ2. Specifically, based on an ad-hoc power analysis, we determined that an AUC difference of 0.05 can be reliably detected using DeLong's test~\cite{delong1988comparing, sun2014fast} by a sample of 1,709 students (> 90\%) but not 591 students (\textasciitilde 40\%). Even though some data augmentation techniques help deal with small sample sizes, the small sample size could bias the initial replicated model for which we need to compare the augmented dataset against. Lastly, the class distribution of the selected target variable is slightly imbalanced: 1,097 students (64\%) enrolled in college (class 1), while the remaining 612 (36\%) did not (class 0). 

\subsubsection{Predictor Variables}
To predict enrollment, two types of features were used: \textit{cognitive} and \textit{affective}, which are both theoretically and empirically supported as influencing student outcomes~\cite{almeda2020predicting,d2012dynamics}. 
The \textit{cognitive} features were students' skill estimates within the 12 topics in mathematics~\cite[Table 2]{zambrano2024long}. To estimate learners' mastery for each topic, Bayesian Knowledge Tracing (BKT) was used~\cite{Pelanek2017}. BKT is a widely recognized technique for tracing learners' knowledge and predicting their knowledge states using data from multiple learning tasks or assessments~\cite{am2021literature}. Employing a commonly-used algorithm that demonstrated satisfactory performance across a wide range of dataset sizes and learning contexts~\cite{slater2018degree}, BKT estimates a learner's current state of knowledge for a specific topic over time, using performance data generated during the interaction with a digital learning platform. Specifically, BKT estimates the probability (a real number between 0 and 1) of a learner correctly applying an underlying skill (e.g., division) to a problem without any instructional support. 
The \textit{affective} features were six measures of students' emotions or behavior: boredom, concentration, confusion, frustration, off-task behavior, and gaming the system. These features align with the model of affective dynamics~\cite{d2012dynamics} and have been commonly used to understand and predict student outcomes~\cite{kai2015comparison}. Due to the space limitations of this paper, further details are available in our code documentation (see link in \Cref{sec:conclusion}).

\subsubsection{Prediction Models}
The original study~\cite{zambrano2024long} employed SML using four different binary classification algorithms: Logistic Regression (LR), Support Vector Machine (SVM), Random Forest (RF), and Multi-Layer Perceptron (MLP). Their implementations come from the open-source Python library \texttt{scikit-learn}~\cite{scikit-learn}, and the original study~\cite{zambrano2024long} used them with default hyperparameters. The metric used to evaluate the models' classification performance is the Area Under the Receiver Operating Curve (AUC)~\cite{Hanley1982auc}, which has been commonly used at LAK~\cite{Liu2024, Jiang2024}. It describes the probability of a classifier correctly distinguishing a positive and negative outcome observation. For each of the four models, the original reported AUC was approximately 0.69~\cite[Table 6]{zambrano2024long}, which is a somewhat satisfactory performance.

\subsection{RQ1: Paper Replication Process}
\label{subsec:methods-rq1}

Two co-authors of this paper used the ASSISTments dataset~\cite{Patikorn2020} and the descriptions in the original study~\cite{zambrano2024long} to implement Python code for replicating the predictive models. A third co-author performed independent testing and code review. Since the original paper does not have the source code available, we implemented ours from scratch. Nevertheless, we later received a private copy of the original code after contacting the authors, so we used this original code to validate our approach. Our version of the implementation is publicly available (see \Cref{sec:conclusion}).

The original study~\cite[Section 3]{zambrano2024long} described most modeling steps, eliminating the need to make arbitrary decisions on our side. After data preprocessing, we verified that the cognitive features matched the original paper within a small error margin. The overall mean absolute difference between the feature values in the paper~\cite[Table 2]{zambrano2024long} and ours was only 0.013. Since the standard deviation of the expected parameter distribution was 0.135, and 0.134 for our computed values, the difference of 0.013 is negligible. The affective features were fully equivalent to the original paper.

The original study used only two classification models (LR and SVM) that are deterministic, meaning that results and model fits are equivalent for each training run of the model. We empirically confirmed that both indeed behave so. 
In contrast, the other two models (RF and MLP) incorporate randomness, and the original study did not use a fixed seed. This would make it difficult to estimate whether our model performances are better or worse by chance, and to what extent they fluctuate by chance.
Therefore, to account for the performance variations, we trained each non-deterministic model 100 times. This iteration count was chosen given the computation time and resource constraints. Subsequently, to ensure a fair performance comparison, we report the average AUCs across the 100 runs (which will also be our baseline for comparing augmentation results). The relevant metric that is averaged across runs is the average AUC across folds, following \citet{zambrano2024long}.
Moreover, to ensure the reproducibility of our results, we set fixed, arbitrary random seeds, which differ for each run. Due to using various seeds and iterating 100 times, this setting ensures that the results of the non-deterministic models are more stable from a random sampling perspective.

\subsection{RQ2: Data Augmentation Process}
\label{subsec:methods-rq2}

After the replication, we assessed numerous classical and contemporary augmentation techniques representing different architectural complexity, ensuring reproducibility by using off-the-shelf libraries. We compared each technique's results to a baseline without augmentation to determine which is the most effective for the given prediction task. Based on related work reviewed in \Cref{subsec:related-work-rq2}, we implemented 21 techniques grouped into 3 categories described in \Cref{table:augmentation_methods}.

\begin{table*}[!ht]
\centering
\caption{Overview of 3 categories of data augmentation methods (we used 21 techniques total) and their LA applications.}

\small
\begin{tabular}{|p{0.105\linewidth}|p{0.63\linewidth}|p{0.10\linewidth}|}
\hline
\textbf{Aug. method} & \textbf{Description and examples} (see \Cref{subsec:related-work-rq2} for details) & \textbf{References} \\ \hline

\textit{Sampling} (9~techniques) & 
Oversampling duplicates minority class instances; undersampling reduces majority class instances. SMOTE generates synthetic minority samples by interpolation, while ADASYN focuses on sparse regions for more targeted oversampling. & 
\cite{chawla2002smote, he2008adasyn, ashfaq2020managing} \\ \hline

\textit{Perturbation} (9~techniques) & 
Small controlled changes to data, like noise injection, to increase variance and prevent overfitting. \citet{wei2019eda} proposed random insertion, swapping, and deletion for text data, which improved performance on small datasets. & 
\cite{wei2019eda, shen2022data} \\ \hline

\textit{Generation} (3~techniques) & 
Generative Adversarial Networks (GANs) create synthetic data by learning the sample distribution, which is particularly useful for imbalanced datasets. GANs have been used in LA to improve knowledge assessment. & 
\cite{berg2016role, biswas2023generative, zhang2024_3dg} \\ \hline
\end{tabular}

\label{table:augmentation_methods}
\end{table*}

We tested all 21 techniques for each of the four models. For each technique, we augmented the input dataset and executed the same modeling code as in RQ1, with a single modification: removing the forward feature selection (FFS) present in the original study and during replication. The reason was that during test runs, it more than quadrupled the execution time, making the convergence of generation methods almost infeasible. Other than that, using the same approach enabled us to assess if augmentation is effective for some model architectures but not others. 

Due to numerous execution combinations (see below), we employed three hardware configurations in parallel:
\begin{itemize}
    \item[\laptop] Laptop: 13th Gen Intel(R) Core(TM) i7-1360P, 2.20 GHz. 16 GB RAM.
    \item[\workstation] Workstation: 13th Gen Intel(R) Core(TM) i9-13900K, 3.00 GHz. 128 GB RAM.
    \item[\cluster] Cluster: Per each task, 1 computing server node out of 150 was used, each with 4 CPU cores and 128 GB RAM.
\end{itemize}

First, we executed each of the 21 augmentation techniques separately. Most of their hyperparameters were set to default (see the code for details). For all three deep generation techniques, we set the hyperparameters to equal values to achieve fair comparison: latent dimension = number of epochs = 100, number of batches = 64. Ideally, we would have used 1000 epochs, but even with the 100 on the stronger hardware, the runtime still takes several hours (see \Cref{table:results}).

Second, we combined pairs of augmentation techniques by sequentially chaining their execution, which has improved prediction results in recent SML research~\cite{Bayer2022, sharma2022smotified}. For the 21 techniques, assuming it does not make sense to apply the same technique twice, there are $21 \times 20 = 420$ possibilities to vary their ordering. However, evaluating 420 chained techniques for each model would not be feasible given long runtimes. Fortunately, the literature established precedents for preferring certain combinations: chaining sampling and then generation~\cite{sharma2022smotified}, or chaining perturbation and then sampling~\cite{Josey2015}. We also chained perturbation and then generation. For the chaining involving generation, we only used the GAN technique, as it is the fastest. Based on \Cref{table:augmentation_methods}, this gives us $9 \times 1 + 9 \times 9 + 9 \times 1 = 99$ combinations to evaluate, each executed again in 100 iterations for each model. To the best of our knowledge, no LA paper has attempted this kind of composite and systematic evaluation of augmentation techniques.

The number of data points affected by augmentation differs for each category of techniques.
For sampling, the minority class data points were boosted to achieve a number of observations equal to the majority class.
For perturbation, no new observations were generated and all existing data points were affected by the transformations.
For deep generation methods, the size of the dataset was doubled.

Finally, after obtaining the results of all augmented models, we assessed which augmentation techniques significantly improved performance. DeLong's test~\cite{delong1988comparing} is the most common method for computing significance between two AUC ROC curves, but it requires predicted class probabilities of a specific model run. Here, we aim to adjust for performance differences that are by chance across several runs, either through specific data points in small samples or random initialization variation in model weights (i.e., seed variation). Therefore, we bootstrapped AUC values within augmentation and baseline methods, meaning we sampled data with replacement and computed the AUC of each sample to induce a distribution of AUC values. These distributions are used to compute empirical distributions of AUC values per model architecture and augmentation method. The empirical distributions are then compared using a $z$-transformation to compute critical quantiles and corresponding $p$-values representing whether predictive methods produce AUC distributions that are significantly different from one another. Given the large number of models compared using $p$-values, we adjusted $\alpha$ for multiple comparisons using the Benjamini-Hochberg correction in R~\cite{Rpackage}, setting the false discovery rate to the conventional 0.05 threshold \cite{benjamini1995controlling}. 

\section{Results and Their Discussion}
\label{sec:results}

\Cref{table:results} provides all results for both RQs. We discuss them separately below. As a reminder, for each row in \Cref{table:results}, we ran non-deterministic models 100 times to mitigate the effects of randomness. We report the mean and standard deviation of the performance measures across the 100 runs, since it is a common practice in LA research~\cite{zambrano2024long, Zhan2023}.

\definecolor{gray}{RGB}{111, 111, 111}

\definecolor{yellow}{RGB}{255, 248, 227}

\definecolor{green}{RGB}{174, 213, 129}
\definecolor{lgreen}{RGB}{220, 237, 200}
\definecolor{red}{RGB}{239, 154, 154}
\definecolor{lred}{RGB}{255, 205, 210}

\newcommand{\gt}[1]{\textcolor{gray}{#1}}  

\newcommand{\gc}{\cellcolor{green}}  
\newcommand{\lgc}{\cellcolor{lgreen}}  
\newcommand{\rc}{\cellcolor{red}}  
\newcommand{\lrc}{\cellcolor{lred}}  

\newcommand{\hlc}[2][yellow]{{%
    \colorlet{foo}{#1}%
    \sethlcolor{foo}\hl{#2}}%
}

\begin{table*}[t]
\centering
\caption{AUC results: mean \gt{(and SD in parentheses)} across the 100 runs. 
For each augmentation category, the best result in each column is highlighted in \hlc[green]{green} (in case of tied mean AUCs, the lower SD wins, the other tied results are \hlc[lgreen]{light green}). 
The worst result in each column is \hlc[red]{red} (in case of tied mean AUCs, the higher SD wins, the other tied results are \hlc[lred]{light red}). 
Next, the bold entries are better than the baseline by at least 0.01. 
Entries prefixed with an asterisk (*) denote that the AUC is statistically significantly different from the baseline after adjusting for multiple testing.
The symbols in the Runtime column refer to the hardware used (see \Cref{subsec:methods-rq2}).
}

\footnotesize
\begin{tabular}{|l|l|cc|cc|c|c|}

\hline

\textbf{Augment.} & \textbf{Method} (FFS = forward & \multicolumn{2}{c|}{\textbf{Deterministic models}} & \multicolumn{2}{c|}{\textbf{Non-deterministic models}} & \textbf{Overall} & \textbf{Runtime}\\[-1mm]
\textbf{category} & feature selection) & \textbf{LR} & \textbf{SVM} & \textbf{RF} & \textbf{MLP} & \textbf{mean} & \textbf{(hh:mm:ss)} \\ \hline

None & Original (with FFS)~\cite{zambrano2024long} & 0.693 \gt{(0.050)} & 0.691 \gt{(0.046)} & 0.692 \gt{(0.037)} & 0.686 \gt{(0.041)} & 0.691 \gt{(0.044)} & N/A \\
None & Replication (with FFS)                      & 0.686 \gt{(0.047)} & 0.652 \gt{(0.046)} & 0.654 \gt{(0.041)} & 0.628 \gt{(0.032)} & 0.655 \gt{(0.041)} & 03:45:16 \ \laptop \\
None & Comparison baseline                         & 0.686 \gt{(0.047)} & 0.652 \gt{(0.046)} & 0.654 \gt{(0.042)} & 0.627 \gt{(0.032)} & 0.655 \gt{(0.042)} & 00:48:54 \ \laptop \\ \hline 

Sampling & SMOTE Standard      & 0.684 \gt{(0.050)}     & 0.659 \gt{(0.049)}              & 0.646 \gt{(0.046)}              & 0.611 \gt{(0.038)}               & 0.650 \gt{(0.046)}              & 00:51:43 \ \laptop \\
Sampling & ADASYN              & \gc 0.688 \gt{(0.048)} & \textbf{0.663} \gt{(0.040)}     & 0.638 \gt{(0.042)}              & 0.608 \gt{(0.033)}               & 0.649 \gt{(0.041)}              & 00:43:23 \ \laptop \\
Sampling & BorderlineSMOTE     & 0.687 \gt{(0.048)}     & \textbf{0.663} \gt{(0.041)}     & 0.640 \gt{(0.040)}              & 0.607 \gt{(0.037)}               & 0.649 \gt{(0.042)}              & \rc 01:16:17 \ \laptop \\
Sampling & KMeansSMOTE         & 0.679 \gt{(0.049)}     & 0.650 \gt{(0.044)}              & 0.657 \gt{(0.043)}              & 0.628 \gt{(0.038)}               & 0.654 \gt{(0.043)}              & 00:57:02 \ \laptop \\
Sampling & SMOTE-Tomek         & 0.685 \gt{(0.051)}     & 0.660 \gt{(0.048)}              & 0.648 \gt{(0.044)}              & 0.618 \gt{(0.037)}               & 0.653 \gt{(0.045)}              & 00:49:41 \ \laptop \\
Sampling & SMOTE-ENN           & 0.681 \gt{(0.062)}     & 0.653 \gt{(0.055)}              & \gc \textbf{0.665} \gt{(0.058)} & \gc *\textbf{0.662} \gt{(0.054)} & \gc \textbf{0.665} \gt{(0.057)} & \gc 00:27:04 \ \laptop \\
Sampling & Random Oversampler  & 0.684 \gt{(0.050)}     & \textbf{0.666} \gt{(0.043)}     & 0.645 \gt{(0.042)}              & 0.607 \gt{(0.043)}               & 0.651 \gt{(0.045)}              & 01:15:33 \ \laptop \\
Sampling & Random Undersampler & 0.687 \gt{(0.046)}     & \gc \textbf{0.676} \gt{(0.038)} & 0.649 \gt{(0.036)}              & 0.632 \gt{(0.033)}               & 0.661 \gt{(0.038)}              & 01:00:11 \ \laptop \\
Sampling & NearMiss            & \rc 0.644 \gt{(0.026)} & \rc *0.581 \gt{(0.029)}         & \rc *0.587 \gt{(0.024)}         & \rc *0.540 \gt{(0.030)}          & \rc 0.588 \gt{(0.027)}          & 00:58:41 \ \laptop \\ \hline 

Perturbation & Polynomial Features  & \gc 0.689 \gt{(0.043)}  & 0.654 \gt{(0.048)}     & 0.653 \gt{(0.037)}     & 0.593 \gt{(0.029)}      & 0.647 \gt{(0.039)}      & \rc 02:00:11 \ \laptop \\
Perturbation & Feature Interaction  & \lrc 0.686 \gt{(0.044)} & 0.653 \gt{(0.047)}     & 0.654 \gt{(0.036)}     & *0.595 \gt{(0.028)}     & 0.647 \gt{(0.038)}      & 01:55:24 \ \laptop \\
Perturbation & PCA                  & \lrc 0.686 \gt{(0.047)} & 0.657 \gt{(0.046)}     & 0.653 \gt{(0.038)}     & *0.593 \gt{(0.026)}     & 0.647 \gt{(0.039)}      & 01:29:40 \ \laptop \\
Perturbation & Standardization      & \lrc 0.686 \gt{(0.045)} & \rc 0.636 \gt{(0.039)} & \rc 0.650 \gt{(0.041)} & *0.574 \gt{(0.033)}     & \lrc 0.637 \gt{(0.040)} & 01:41:28 \ \laptop \\
Perturbation & Min-Max Scaling      & \lrc 0.686 \gt{(0.045)} & 0.652 \gt{(0.039)}     & 0.651 \gt{(0.041)}     & 0.607 \gt{(0.034)}      & 0.649 \gt{(0.040)}      & 01:11:39 \ \laptop \\
Perturbation & Robust Scaling       & \lrc 0.686 \gt{(0.045)} & \gc 0.661 \gt{(0.037)} & 0.651 \gt{(0.041)}     & *0.583 \gt{(0.033)}     & 0.645 \gt{(0.039)}      & 01:28:54 \ \laptop \\
Perturbation & Log Transformation   & \lrc 0.686 \gt{(0.046)} & 0.652 \gt{(0.045)}     & 0.651 \gt{(0.041)}     & 0.622 \gt{(0.034)}      & 0.653 \gt{(0.041)}      & 01:03:10 \ \laptop \\
Perturbation & Power Transformation & \lgc 0.689 \gt{(0.046)} & 0.639 \gt{(0.047)}     & 0.652 \gt{(0.040)}     & \rc *0.569 \gt{(0.030)} & \rc 0.637 \gt{(0.041)}  & 01:42:53 \ \laptop \\
Perturbation & Noise Addition       & \rc 0.686 \gt{(0.048)}  & 0.652 \gt{(0.044)}     & \gc 0.655 \gt{(0.041)} & \gc 0.629 \gt{(0.033)}  & \gc 0.656 \gt{(0.042)}  & \gc 01:02:08 \ \laptop \\ \hline

Generation & GAN  & \gc 0.686 \gt{(0.048)} & 0.648 \gt{(0.046)}              & 0.653 \gt{(0.039)}     & \textbf{0.641} \gt{(0.034)}     & \gc 0.657 \gt{(0.042)}  & \gc 02:26:18 \ \workstation \\
Generation & VAE  & 0.682 \gt{(0.047)}     & \rc 0.637 \gt{(0.043)}          & \rc 0.652 \gt{(0.037)} & \rc 0.635 \gt{(0.036)}          & \lrc 0.652 \gt{(0.041)} & 12:07:27 \ \workstation \\
Generation & CGAN & \rc 0.640 \gt{(0.050)} & \gc \textbf{0.665} \gt{(0.044)} & \gc 0.656 \gt{(0.043)} & \gc \textbf{0.649} \gt{(0.033)} & \rc 0.652 \gt{(0.042)}  & \rc 28:52:40 \ \workstation \\ \hline

Best S + G & SMOTE-ENN + GAN            & \rc 0.682 \gt{(0.060)} & 0.655 \gt{(0.057)}     & 0.663 \gt{(0.056)}              & \textbf{0.662} \gt{(0.055)}      & \textbf{0.666} \gt{(0.057)}     & 16:17:55 \ \workstation \\ 
Best P + S & Noise Addition + SMOTE-ENN & 0.684 \gt{(0.056)}     & \gc 0.658 \gt{(0.051)} & \gc \textbf{0.665} \gt{(0.055)} & \gc *\textbf{0.667} \gt{(0.054)} & \gc \textbf{0.668} \gt{(0.054)} & 00:39:14 \ \laptop \\ 
Best P + G & Noise Addition + GAN       & \gc 0.687 \gt{(0.048)} & \rc 0.648 \gt{(0.047)} & \rc 0.655 \gt{(0.038)}          & \rc \textbf{0.641} \gt{(0.036)}  & \rc 0.657 \gt{(0.042)}          & 01:08:15 \ \cluster \\ \hline

\end{tabular}

\label{table:results}
\end{table*}

\subsection{RQ1: Paper Replication}

Based on the first two rows of \Cref{table:results}, we replicated similar model performances as in the original paper~\cite{zambrano2024long}. The total mean AUC difference between the original and our models is 0.036, which is minor. Moreover, the difference between the original best model and our best model (both LR) is just 0.007. Zambrano and Baker's \cite{zambrano2024long} results are within the 95\% confidence interval of our results, which can be considered a successful replication within the uncertainty of seed variation not accounted for in the original paper. Another evidence indicating close replication is that the relative order of the models' performance is preserved with respect to the original study (LR > RF > SVM > MLP). It is interesting that in both original and our study, the simplest model architecture performed the best, while the most complex model performed the worst. The observed performance differences can be attributed to the random variation caused by the unknown seed in the original paper. Seed variation is a common factor preventing exact reproducibility at LAK~\cite{Haim2023lak}.

Additionally, our results are consistent with previous research, which has demonstrated that both cognitive and affective factors are predictive of college enrollment~\cite{pedro2013predicting,san2022exploring}. These findings also reinforce the central hypothesis of the model of affective dynamics~\cite{d2012dynamics}, which posits that emotional states and knowledge-related factors (i.e., mathematical skill estimates in this case) over time have a significant impact on long-term academic outcomes.

\subsection{RQ2: Data Augmentation}

\Cref{table:results} also compares our baseline to our augmentation results. Since the replication results are almost exactly the same regardless of whether FFS is used or not, we use the version without FFS as the baseline, because removing FFS substantially reduced the runtime (to about 22\% of the original).

\subsubsection{Sampling}
Overall, \textit{SMOTE-ENN} is the best performing sampling technique, improving the overall mean AUC by 0.01 across model architectures and even reducing the training time to just 55\% of the baseline. The improvement is higher for the non-deterministic models, especially MLP (+0.035), where it is also statistically significant. On the contrary, \textit{NearMiss} is the worst technique, reducing all four model performances (significantly in three models) while also prolonging the training by 20\%. 
Among the individual models, LR is barely influenced by any technique (except for the decrease with \textit{NearMiss}). For SVM, \textit{Random Undersampler} (+0.025) and \textit{Random Oversampler} (+0.014) are beneficial.

\subsubsection{Perturbation}
When applied in isolation, perturbation did not boost model performance. For LR and RF, it had almost no effect. For MLP, the result was almost always worse, often significantly. For SVM, the improvements were smaller than 0.01. Similarly negligible improvement was observed for RF and MLP with \textit{Noise Addition}. However, in all cases, the training time also increased by at least 27\%, so perturbation alone was not useful in this prediction task. 

\subsubsection{Generation}
Regarding the deep generation techniques, which are the most complex category, even the best result for each column was always outperformed by at least one simpler sampling technique. In other words, some sampling technique was better than the best generation technique for all four ML model architectures. This is an important finding because the sampling techniques require a fraction of the training time. When compared to perturbation, GAN and CGAN improved the MLP model (+0.014 and +0.022, respectively, compared to the baseline). However, CGAN also notably worsened the LR model, showing that not all techniques are suitable for all model architectures.

\subsubsection{Technique Combinations}
For the combined augmentation techniques, since the number of combinations is so large, we only report the best combinations per category. The precedents established in the literature (sampling and then generation~\cite{sharma2022smotified}, or perturbation and then sampling~\cite{Josey2015}) yielded only a microscopic improvement over a single technique (AUC increase of less than 0.01). Specifically, both \textit{SMOTE-ENN} and \textit{Noise Addition}, which were alone the overall best in their respective categories (+0.01 and +0.001 improvement over the baseline), did also the best together (+0.013 improvement over the baseline) among all combinations. Overall, the implied finding is that if a technique was not useful alone, then including it in a combination will not make it substantially better. However, if a technique was promising alone, then it is worth trying it in combination with another augmentation category.

\subsection{Limitations}

Compared to related work, a previous LA study~\cite{ashfaq2020managing} reported improved accuracy (not AUC) of SVM, RF, and neural network models when using SMOTE and ADASYN (which our results did not determine to be useful). However, that study used a smaller dataset of 480 students and did not cross-validate on an ID attribute (we used school-level cross-validation~\cite{zambrano2024long}), so their test set performance may have been inflated. Other prior studies that used data augmentation in LA employed different algorithms, so comparing to their results is not possible. Finally, in non-LA domains, the improvements after data augmentation ranged from substantial, to minor, to sometimes even negative~\cite{Bayer2022}.

Deep generation introduced more variation in AUCs across seeds and did not improve performance. These methods may need hyperparameter tuning (e.g., by random search) to be effective, and appear impractical for most LA practitioners whose focus is on model performance. Other settings might make deep generation more effective, but that would require considerable computational resources and time.
Moreover, non-deterministic models combined with deep generation contain other forms of non-determinism besides the seed. We observed that during the technique chaining, even though most iterations finished fine, some could produce null values and crash on an error we did not encounter during testing.

We did not inspect feature selection because it did not change the baseline performance and substantially slowed the model fitting. Yet, feature selection could be explored for improving feature engineering \textit{after} augmentation is applied, especially in the context of larger LA datasets.

Lastly, our results are limited to the replicated paper's context. Expanding the evaluation could reveal more general properties of augmentation, allowing the findings to be reliably generalized to another LA context.

\subsection{Practical Observations and Implications for the Field of LA}

Our results bear several implications for LA researchers and practitioners who seek to use data augmentation in predictive modeling.
First, we observed considerable variation in how much different augmentation techniques improved -- or, at times, worsened -- the performance of different models. If researchers have limited time, certain techniques, such as SMOTE-ENN, might be more worth trying than others.
Second, we note training runtime tradeoffs. Most performance improvements come at the cost of longer training, but, surprisingly, some techniques can both improve performance and decrease runtime.
Third, often neglected in past research, it seems challenging to establish statistical significance when testing several augmentation techniques on sample sizes typical for our field. We provide open-source code (see \Cref{sec:conclusion}) to facilitate rigorous significance testing for this purpose.

Our study provided new insights indicating that for binary classification, most augmentation methods are too complex to yield performance improvements without extensive hyperparameter tuning (which may be impractical given the runtimes noted here). However, sampling can slightly improve model performance, as observed with SMOTE-ENN, particularly on the non-deterministic models. Researchers can easily use this method for predictive modeling, as SMOTE-ENN is implemented in the popular Python library \texttt{scikit-learn}~\cite{scikit-learn}. It is important to note that other researchers may not need to run 100 iterations; we did this only to identify the most effective methods in a statistically reliable manner. For practical applications, one iteration may suffice, converging within minutes on standard PCs.

From a technical perspective, SMOTE-ENN involves two key steps~\cite{Nishat2022smoteenn}: after SMOTE generates synthetic samples by interpolating between existing observations (i.e., oversampling), ENN helps clean up the dataset (i.e., regularization by noise reduction). It does this by removing instances (both original and synthetic) that are misclassified by their nearest neighbors. This step reduces noisy instances, particularly near the class boundaries, thus refining the data. Notably, SMOTE-ENN was further boosted by additionally combining it with noise addition. While, at first glance, the two methods seem to have contradictory objectives, we interpret this boost as noise addition further regularizing the dataset, boosting performance on a holdout fold during cross-validation while preventing overfitting.

This is a likely reason why this process contributed to improving the performance, given that some signal in the original dataset is noise. All the features used in the models (both cognitive and affective) only estimate real-world constructs and thus have inaccuracies, as was reported in a recent paper~\cite{Liu2024affect} that used a similar dataset from the same digital learning platform, ASSISTments.

Overall, our findings have implications for binary classification in LA and other prediction tasks involving SML, which are prevalent in educational data science. We also used cognitive and affective features that are common predictors in other LA studies. Therefore, researchers operating in similar contexts can try using data augmentation with SMOTE-ENN sampling, in order to increase the overall quality of training data and decrease the training time. 

\subsection{Open Research Challenges}

Future work should evaluate the methods on other datasets and prediction tasks (beyond enrollment outcomes and binary classification). Studying other model architectures and hyperparameter settings is also warranted. We generally observed that simpler models did better, potentially because their complexity regularizes prediction more. Hence, future work may study if regularization of more complex neural networks, for example through dropout, may further boost the performance gains related to data augmentation reported here.

Another direction is to output the augmentation results, visualize how the data distribution changed after the augmentation, and examine the statistical properties of the augmented data samples. Subsequently, having such dataset allows us to directly investigate why some of the augmentation techniques work and why others do not. This includes their impact on feature importance, which we did not consider in this study.

Such research can also vary the amount of data produced by augmentation. So far, the most successful techniques in this study (sampling) produced new data samples to ensure balanced classes (i.e., up to equal class size). Our findings merit further study comparing the differences in results if, after augmentation, the size of one class exceeded the other by a certain ratio, such as $2\times$ or $3\times$ the size.
In addition, by experimentally controlling the number of augmented data points, we can study what the most desirable ratio of synthetic to real data in LA prediction tasks is.

Researchers interested in generative models like large language models can employ them to generate additional training data, enabling learning that extends beyond the patterns present in the existing data. Related to this, future work can examine how datasets with different degrees of similarity contribute to predictive improvements. This involves measuring similarities between the original and various augmented datasets to determine how different the augmented data are and whether more different or more similar datasets improve predictions.

\section{Conclusion}
\label{sec:conclusion}

Using data augmentation in LA contexts is a timely topic. Much of LA research employs supervised machine learning, often using small or imbalanced data (e.g., for tasks such as dropout prediction). Within other domains, data augmentation often improved predictive performance. Yet, to the best of our knowledge, no LA paper has attempted a systematic empirical evaluation of data augmentation for outcome prediction. 

Our study assessed 21 data augmentation techniques, including their 99 chained combinations. We observed that traditional sampling techniques are as good or better than more recent, advanced techniques, which take much longer to compute and are less stable due to non-determinism and more hyperparameters. Further, chaining yielded tangible, although small improvements beyond sampling alone. Lastly, the most effective technique, SMOTE-ENN, which significantly improved predictive performance, was also the fastest to run (even much faster than not performing any data augmentation). These results suggest that off-the-shelf sampling techniques may provide the most applicable improvements in LA contexts. Still, some transformation and generation techniques worsened predictive performance, so careful method selection is necessary.

Although the overall AUC improvements may seem small, sometimes even small improvements in predictive performance matter. The practical relevance of these improvements depends on context and application (e.g., scale, risk, and benefits of correct classification) and is beyond the scope of this paper. Our contribution includes open-source materials for other researchers to conduct rigorous significance testing of improvements resulting from a suite of augmentation techniques. Ultimately, our research tested the utility of data augmentation in LA, providing insights into its applicability and the conditions under which it may or may not enhance predictive performance.

In parallel, our second contribution was independently validating the results from a previous LAK paper~\cite{zambrano2024long}. By independently reproducing results using our own analysis code based on the original study, we demonstrated further evidence to the LA community regarding the robustness of the study's findings. From a broader perspective, replication is a cornerstone of scientific progress. Our results serve to reinforce the earlier work, which is rare in LA research, and contribute towards addressing the reproducibility crisis~\cite{Hutson2018crisis, Haim2023lak, Haim2023edm}. 

All code used in our research and full results are available~\cite{paper-supplementary-materials}. The dataset must be requested from its owners~\cite{Patikorn2020}.

\begin{acks}
This work was supported by JST CREST Grant Number JPMJCR22D1, JSPS KAKENHI Grant Number JP22H00551, and MEXT \enquote{Innovation Platform for Society 5.0} Program Grant Number JPMXP0518071489, Japan.
We thank Andres Felipe Zambrano for providing the code for the original study.
This work benefited from the resources of the Institute of Advanced Computing at Tampere Center for Scientific Computing.
\end{acks}

\bibliographystyle{ACM-Reference-Format}
\bibliography{references}

\end{document}